\documentclass[nohyperref]{article}

\usepackage{microtype}
\usepackage{multirow}
\usepackage{graphicx}
\usepackage{subfigure}
\usepackage{booktabs} % for professional tables

\usepackage{hyperref}

\usepackage[accepted]{icml2022}

\usepackage{amsmath,amsfonts,bm}

\def\eqref#1{equation~\ref{#1}}
\def\1{\bm{1}}

\def\vtheta{{\bm{\theta}}}

\def\vf{{\bm{f}}}

\def\vx{{\bm{x}}}

\def\mF{{\bm{F}}}

\def\mP{{\bm{P}}}

\def\mU{{\bm{U}}}
\def\mV{{\bm{V}}}
\def\mW{{\bm{W}}}
\def\mX{{\bm{X}}}

\DeclareMathAlphabet{\mathsfit}{\encodingdefault}{\sfdefault}{m}{sl}
\SetMathAlphabet{\mathsfit}{bold}{\encodingdefault}{\sfdefault}{bx}{n}

\newcommand{\E}{\mathbb{E}}

\newcommand{\R}{\mathbb{R}}

\usepackage{amsmath}
\usepackage{amssymb}
\usepackage{mathtools}
\usepackage{amsthm}

\usepackage[capitalize,noabbrev]{cleveref}

\theoremstyle{plain}

\theoremstyle{definition}

\theoremstyle{remark}

\usepackage[textsize=tiny]{todonotes}

\icmltitlerunning{Unsupervised Time-Series Representation Learning with Iterative Bilinear Temporal-Spectral Fusion}

\begin{document}

\twocolumn[
\icmltitle{Unsupervised Time-Series Representation Learning with \\Iterative Bilinear Temporal-Spectral Fusion}




\begin{icmlauthorlist}
\icmlauthor{Ling Yang}{yyy,sch}
\icmlauthor{Shenda Hong}{yyy,sch}
% \icmlauthor{Firstname3 Lastname3}{comp}
\end{icmlauthorlist}

\icmlaffiliation{yyy}{National Institute of Health Data Science, Peking University, Beijing, China}
\icmlaffiliation{sch}{Institute of Medical Technology, Health Science Center of Peking University, Beijing, China}

\icmlcorrespondingauthor{Ling Yang}{yangling0818@163.com}
\icmlcorrespondingauthor{Shenda Hong}{hongshenda@pku.edu.cn}

\icmlkeywords{Machine Learning, ICML}

\vskip 0.3in
]



\printAffiliationsAndNotice{} % leave blank if no need to mention equal contribution

\begin{abstract}
Unsupervised/self-supervised time series representation learning is a challenging problem because of its complex dynamics and sparse annotations. Existing works mainly adopt the framework of contrastive learning with the time-based augmentation techniques to sample positives and negatives for contrastive training. Nevertheless, they mostly use segment-level augmentation derived from time slicing, which may bring about sampling bias and incorrect optimization with false negatives due to the loss of global context. Besides, they all pay no attention to incorporate the spectral information in feature representation. In this paper, we propose a unified framework, namely Bilinear Temporal-Spectral Fusion (\textit{BTSF}).
Specifically, we firstly utilize the instance-level augmentation with a simple dropout on the entire time series for maximally capturing long-term dependencies. We devise a novel \textit{iterative bilinear temporal-spectral fusion} to explicitly encode the affinities of abundant time-frequency pairs, and iteratively refines representations in a fusion-and-squeeze manner with Spectrum-to-Time (S2T) and Time-to-Spectrum (T2S) Aggregation modules. We firstly conducts downstream evaluations on three major tasks for time series including classification, forecasting and anomaly detection. Experimental results shows that our BTSF consistently significantly outperforms the state-of-the-art methods. 
% Code will be released upon acceptance.
\end{abstract}

\section{Introduction}
\label{intro1}
Time series analysis \citep{Oreshkin2020N-BEATS:} plays a crucial role in various real-world  scenarios, such as traffic \begin{figure}[H]
\vskip 0.1in
\centering
    \includegraphics[width=6.5cm]{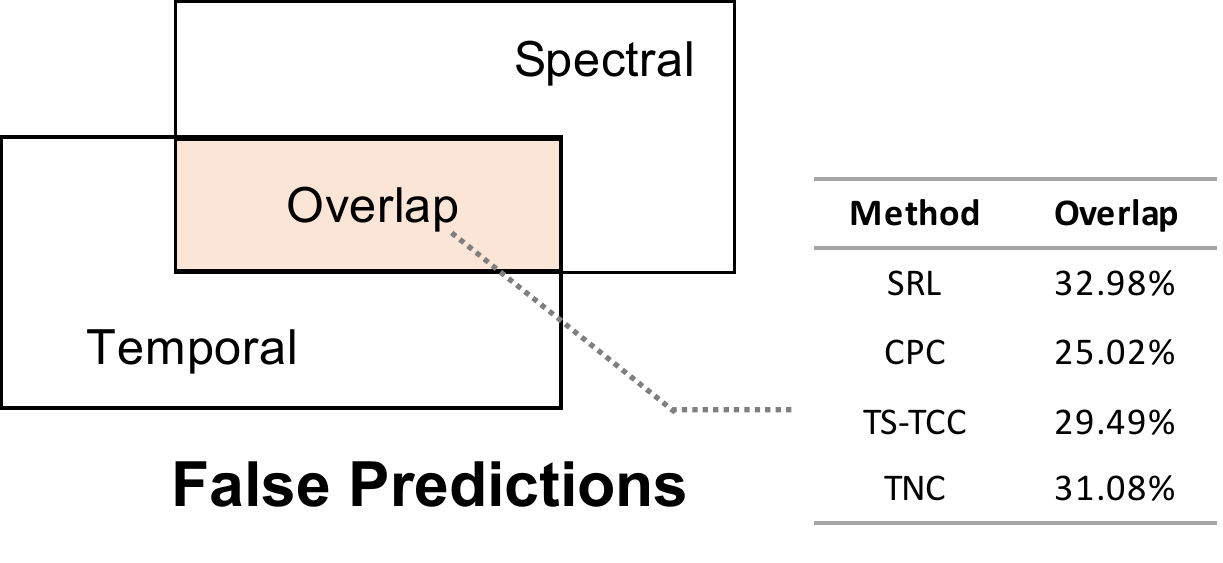}
\caption{Statistics about false predictions of randomly selected evaluation samples.}
\label{statistics}
\end{figure}
prediction, clinical trials and financial market. 
Classification \cite{esling2012time}, forecasting \citep{DEB2017902} and anomaly detection \citep{10.1145/2783258.2788611} are main tasks for time series analysis. However, there is often no adequate labeled data for training and results are not ideal when time series are sparsely labeled or without supervision \citep{lan2021intrainter}. Therefore, it is valuable to study on the unsupervised representation learning for time series with which the learned representations can be used for aforementioned downstream tasks. Unsupervised representation learning has been well studied in computer vision and natural language processing \citep{denton2017unsupervised,gutmann2012noise,wang2015unsupervised,pagliardini2018unsupervised,pmlr-v119-chen20j} but only a few researches are related with time series analysis \citep{eldele2021time,yue2021learning,liu2021spatio}.

\begin{figure*}[h]
\vskip 0.15in
\centering
    \includegraphics[width=13cm]{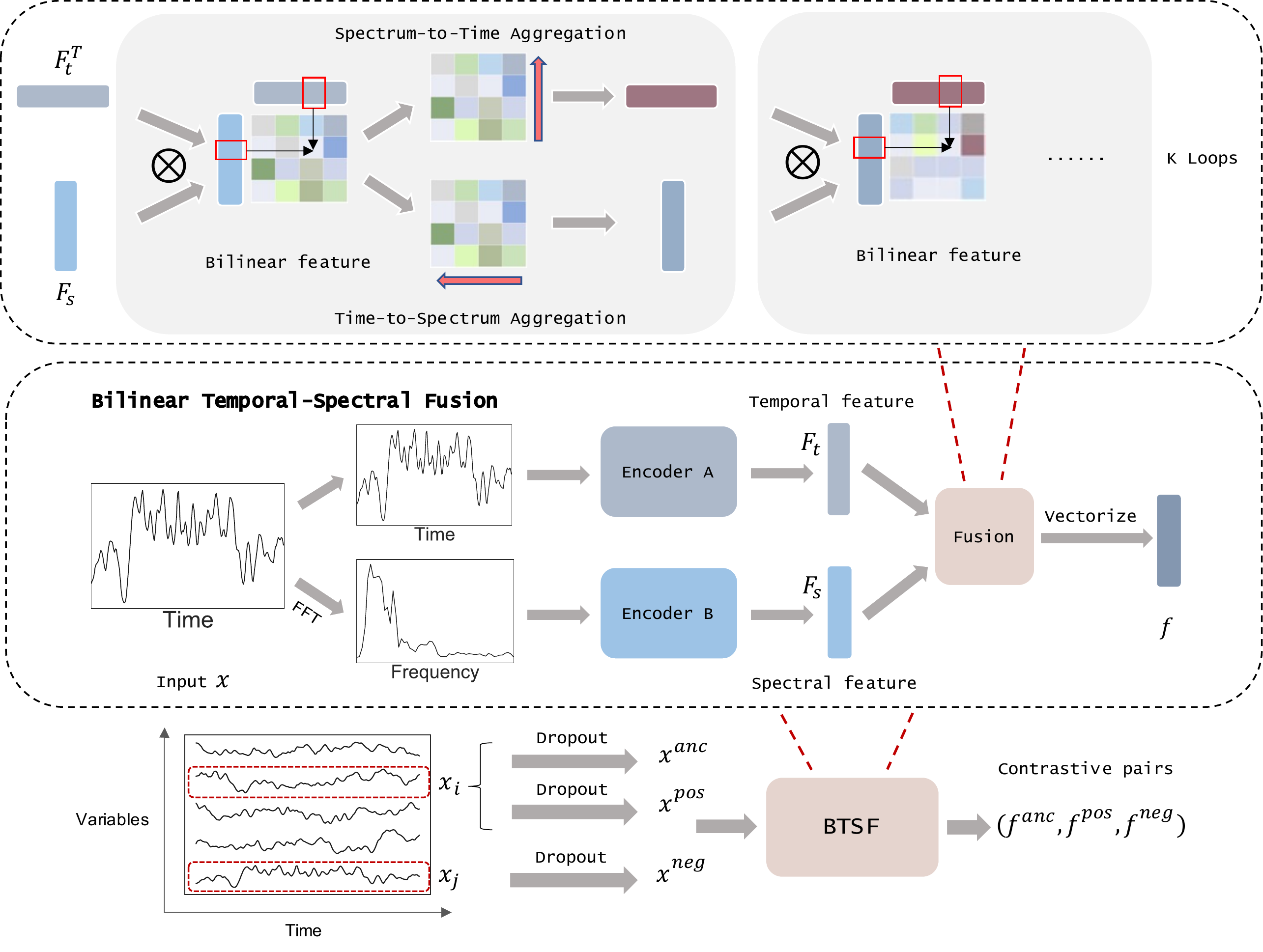}
\caption{The diagram of our general unsupervised representation learning framework for multivariate time series, $\otimes$ is the cross product. See Section \ref{btsf} for more details.}
\label{fig:bilinear}
\vskip 0.1in
\end{figure*}
Recent works mainly utilize the time-based contrastive learning framework \citep{chen2020simple,zerveas2021transformer} for unsupervised representation learning in time series.
Time-Contrastive Learning (TCL) \citep{hyvarinen2016unsupervised}, Contrastive Predictive Coding (CPC) \citep{oord2018representation}, Scalable RepresentationLearning (SRL) \citep{franceschi2019unsupervised}, Temporal and Contextual Contrasting (TS-TCC) \citep{eldele2021time} and Temporal Neighborhood Coding(TNC) \citep{tonekaboni2021unsupervised} are all segment-level methods which sample contrastive pairs along temporal axis. Nevertheless, they all fail to utilize the temporal-spectral affinities in time series and thus limit the discriminativity and expressiveness of the representations. We further take an experimental analysis on these methods and Figure \ref{statistics} shows statistics about false predictions on time series classification.
We implement existing works according to public codes. In specific, \textit{by spectral} means we use their proposed sampling methods to generate contrastive pairs and transform the sampled time series into spectral domain for extracting feature for later training and testing. It is notable that existing works all have a low overlap percentage around 30\% about false predictions with only temporal or spectral feature. The phenomenon demonstrates their temporal and spectral representations have few associations. Besides, previous segment-level methods are based on the assumption that distant segments are negative pairs and neighbour segments are positive pairs, which usually perform badly in long-term scenarios and fail to capture the global context.

Based on the aforementioned shortcomings of existing works, we propose an unsupervised representation learning framework for time series, namely Bilinear Temporal-Spectral Fusion (BTSF). BTSF promotes the representation learning process from two aspects, the more reasonable construction of contrastive pairs and the adequate integration of temporal and spectral information. In order to preserve the global temporal information and have the ability to capture long-term dependencies of time series, BTSF uses the entire time series as input  and simply applies a standard dropout \citep{JMLR:v15:srivastava14a} as an instance-level augmentation to produce different views of time series.
Such construction of contrastive pairs ensures that the augmented time series would not change their raw properties, which effectively reduces the possible false negatives and positives. 
For the effective combination of temporal-spectral information and further achieving alignment between them in feature representation, we perform an iterative bilinear fusion between temporal and spectral features to produce a fine-grained second-order feature which explicitly preserves abundant pairwise temporal-spectral affinities. To adequately utilize the informative affinities, we further design a cross-domain interaction with Spectrum-to-Time and Time-to-Spectrum Aggregation modules to iteratively refine temporal and spectral features for cycle update. Compared to simple combination operations like summation and concatenation, our bilinear fusion make it possible that the temporal (spectral) feature gets straightly enhanced by spectral (temporal) information of the same time series, which is proved to be effective by our further experiments and theoretical analysis.
% Due to the property of our bilinear fusion, BTFS is scalable with no limit on the length of time series. 
% To measures the representation quality, we also make assessments from the perspectives of alignment and uniformity  with the learned representations for time series. In addition, we employ the obtained representations for different downstream tasks to evaluate the superior generalization ability.

Our main contributions are summarized as the following:
\begin{itemize}
    \item We revisit the existing segment-level contrastive learning framework for time series representation learning and propose the instance-level augmentation technique to maximally preserve global context.
    \item A novel \textit{iterative bilinear temporal-spectral fusion} is proposed to explicitly model pairwise cross-domain dependencies for discriminating and enriching representations in a fusion-and-squeeze manner. 
    \item Sufficient assessments including \textit{alignment} and \textit{uniformity} \citep{wang2020understanding} are conducted to identify the generalization ability of our learned representations.
    \item Extensive experiments show that our BTSF significantly outperforms previous works in downstream classification, forecasting and anomaly detection tasks, and is competitive with supervised methods. 
\end{itemize}

\section{Related Work}
\label{related work}
\paragraph{Unsupervised Representation Learning for Time Series.} A relevant direction of research about representation learning on sequence data has been well-studied \citep{chung2015recurrent,fraccaro2016sequential,krishnan2017structured,bayer2021mind}. However, few efforts have made in unsupervised representation learning for time series \citep{langkvist2014review,eldele2021time,yue2021learning}. Applying auto-encoders \citep{choi2016multi} and seq-to-seq models \citep{malhotra2017timenet,lyu2018improving} with an encoder-decoder architecture to reconstruct the input are preliminary approaches to unsupervised representation learning for time series. Rocket \citep{dempster2020rocket} is a fast method that involves training a linear classifier on top of features extracted by a flat collection of numerous and various random convolutional kernels.
% While the construction is more challenging in cases like for high-frequency physiological signals, several approaches are proposed to handle it. 
Several approaches leverage inherent correlations in time series to learn unsupervised representations.
SPIRAL \citep{lei2017similarity} bridges the gap between time series data and static clustering algorithm through preserving the pairwise similarities of the raw time series data. \citet{ma2019learning} integrates the temporal reconstruction and K-means \citep{krishna1999genetic} objective to generate cluster-specific temporal representations.

\paragraph{Time-Series Contrastive Learning.} Another group of approaches design different sample policy and incorporate contrastive learning \citep{hyvarinen2016unsupervised,oord2018representation,chen2020simple,yue2021ts2vec} to tackle representation learning for temporal data without supervision.  
Inspired by Word2Vec \citep{mikolov2013efficient}, Scalable Representation Learning (SRL) \citep{franceschi2019unsupervised} proposes a novel triplet loss and tries to learn scalable representations via randomly sampling time segments. Contrastive Predictive Coding (CPC) \citep{oord2018representation} conducts representation learning by using powerful autoregressive models in latent space to make predictions in the future, relying on Noise-Contrastive Estimation \citep{pmlr-v9-gutmann10a} for the loss function in similar ways. Temporal and Contextual Contrasting (TS-TCC) \citep{eldele2021time} is a improved work of CPC and learns robust representation by a harder prediction task against perturbations introduced by different timestamps and augmentations. Temporal Neighborhood Coding (TNC) \citep{tonekaboni2021unsupervised} presents a novel neighborhood-based unsupervised learning framework and applies sample weight adjustment for non-stationary multivariate time series. Their main difference is that they select contrastive pairs according to different segment-level sampling policies.
% They all use time slicing as augmentation to generate different views for constructing positive and negative pairs in representation learning. 
However, they are prone to be affected by false negatives and fails to capture long-term dependencies because of the loss of global context. Besides, they only extract temporal feature, neglecting to leverage spectral feature and involve temporal-spectral relations. In this paper, we address all these problems in a unified framework.

\section{The Proposed Method}

\subsection{Instance-level Augmentation Technique}
Previous researches on the unsupervised representation learning for time series mainly tackle the problem by designing different sampling policy on temporal data. They use the sampled data to construct the contrastive objective for guiding the training procedure. Sampling bias is an inevitable problem for existing representation works in time series because of their segment-level sampling policy (time slicing). Time slicing is unable to capture the long-term dependencies due to the loss of global semantical information. To explore an effective augmentation method for the construction of contrastive pairs, we first investigate general augmentation methods for time series. A latest empirical survey \citep{iwana2021empirical} evaluates 12 time series data augmentation methods on 128 time series classification datasets with 6 different types of neural networks. According to results, no augmentation method, not excepting time slicing, is able to improve performance on all datasets consistently. It is because time series is sensitive to sequential order and temporal patterns. 

To preserve the global temporal information and not change the original properties for time series, we apply a standard dropout as a minimal data augmentation to generate different views in unsupervised representation learning. Specifically, we simply employ two independently sampled dropout masks on the time series to obtain a positive pair and treat time series of other variables as negative samples for negative pairs construction. With the instance-level contrastive pairs, our method has the ability to capture long-term dependencies and effectively reduce the sampling bias which is superior to previous segment-level pairs. In the procedure of contrastive pairs construction, we pass the each time series $\vx$ to the dropout to generate a positive pair $\vx^{anc}$ and $\vx^{pos}$.
For negative samples, we randomly choose other variables as $\vx^{neg}$ for multivariate time series. 
\begin{align}
\label{enc-eq}
    \vx^{anc} = Dropout(\vx),\quad
    \vx^{pos} = Dropout(\vx).
\end{align}
Thus our instance-level augmentation is general and can process both non-stationary and periodic time series. In contrast, time slicing fails to deal with the periodic time series because it is possible for them to choose false negative samples. The dropout rate is set to 0.1 in our experiments. For more experimental comparisons with other augmentation methods and the sensitivity of dropout rate, see Appendix \ref{more ablation} for more details.
\subsection{Iterative Bilinear Temporal-Spectral Fusion}
\label{btsf}
In this subsection, we provide a detailed introduction to a general and effective framework for learns a discriminative feature representation for multivariate time series, namely Bilinear Temporal-Spectral Fusion (BTSF). As illustrated in Figure \ref{fig:bilinear}, after constructing the contrastive pairs, we map the time series to a high dimensional feature space to assimilate $\vx$ and $\vx^{pos}$, and to distinguish $\vx^{neg}$ from $\vx$. Previous works neglect to
leverage spectral feature and temporal-spectral relations, our proposed BTSF not only simultaneously utilize spectral and temporal features but also enhances the representation learning in a more fine-grained way. Instead of summation and concatenation, BTSF adopts iterative bilinear temporal-spectral fusion to iteratively explore and refine the pairwise affinities between temporal and spectral features for producing an interactive feature representation, representing the most common parts of positive pairs and enlarging the differences of negative pairs.   

Specifically, each augmented time series $\vx_t$ is first transformed to spectral domain by a fast Fourier transform (FFT), obtaining spectral signal $\vx_s$. Then $\vx_t$ and $\vx_s$ are delivered to two encoding networks for feature extraction respectively. The process is as the following:
\begin{align}
\label{enc-eq}
    \mF_t = Encoder_A(\vx_t ; \vtheta_t),\quad
    \mF_s = Encoder_B(\vx_s ; \vtheta_s)
\end{align}
where $\mF_t \in \R^{m\times d}$ and $\mF_s \in \R^{n\times d}$ are temporal and spectral features, $\vtheta_t$ and $\vtheta_s$ are parameters of their encoding networks $Encoder_A$ and $Encoder_B$ respectively. We just use simple stacks of dilated causal convolutions \citep{bai2018empirical} to encode temporal features and use 1D convolutional blocks to extract spectral features. We apply a max-pooling layer in the end of encoding network to guarantee the same size of features, which makes our model scalable to input length. BTSF makes an iterative bilinear fusion between $\mF_t$ and $\mF_s$. Specifically, we establish a channel-wise interaction between features of two domains as the following:
\begin{align}
    \mF(i,j) = {\mF_t(i)}^T\mF_s(j)
\end{align}
where $i$ and $j$ stand for the $i$-th and $j$-th
location in temporal and spectral axes respectively. This bilinear process adequately models the fine-grained time-frequency affinities between $\mF_t(i) \in \R^{d}$ and $\mF_s(i) \in \R^{d}$. To summarize such affinities globally, BTSF integrates $\mF(i,j) \in \R^{d\times d}$ to produce the initial bilinear feature vector $\mF_{bilinear} \in \R^{d\times d}$ with sum pooling of all time-frequency feature pairs:
\begin{align}
\label{ori-eq}
\begin{split}
    \mF_{bilinear} &= {\mF_t}^T\times\mF_s = \sum_{i=1}^{m}\sum_{j=1}^{n}\mF(i,j) \\
    &=\sum_{i=1}^{m}\sum_{j=1}^{n}{\mF_t(i)}^T\mF_s(j)
\end{split}
\end{align}
where $\times$ denotes the matrix multiplication.
This bilinear feature conveys the fine-grained time-frequency affinities to acquire a more discriminative feature representation. Then we encode cross-domain affinities to adaptively refine the temporal and spectral features through an iterative procedure as the following:
\begin{equation}
\label{loops}
\begin{aligned}
    \operatorname{S2T}:\quad \mF_t &= BiCasual(Conv(\mF_{bilinear}))\\
    \operatorname{T2S}:\quad
    \mF_s &= Conv(BiCasual(\mF_{bilinear}))
\end{aligned}
\end{equation}
where $\mF_t \in \R^{m\times d}$ and $\mF_s \in \R^{n\times d}$ are updated by Spectrum-to-Time Aggregation ($\operatorname{S2T}: \R^{d\times d} \rightarrow \R^{m\times d}$) and Time-to-Spectrum Aggregation ($\operatorname{T2S}: \R^{d\times d} \rightarrow \R^{n\times d}$). $Conv$ is normal convolution and $BiCasual$ is bi-directional causal convolution, followed by nonlinear function (e.g., ReLU). Specifically, $\operatorname{S2T}$ first aggregates spectrum-attentive information for each temporal feature through applying convolutional blocks along spectral axis. Then it exchanges the spectrum-related information along temporal axis to refine the temporal features by several bi-directional causal convolutions. Contrary to $\operatorname{S2T}$, $\operatorname{T2S}$ applies above aggregation-exchange procedure from temporal domain to spectral domain. $\operatorname{S2T}$ and $\operatorname{T2S}$ modules adequately aggregate the cross-domain dependencies and refine the temporal and spectral features respectively. In turn, refined temporal and spectral features are able to produce more discriminative bilinear feature. $\operatorname{S2T}$, $\operatorname{T2S}$ and bilinear fusion jointly form a loop block in a fuse-and-squeeze manner. After several loops  of Eq.(\ref{ori-eq}) and Eq.(\ref{loops}), the final bilinear feature $\mF_{bilinear}$ is obtained. The ablation study of loops number is in Appendix \ref{more ablation}.
% \begin{align}
% \label{ori-eq}
%     \vf_{bilinear} = vec({\mF_t}^T\mF_s) =  vec(\sum_{i=1}^{m}\sum_{j=1}^{n}\mF(i,j)) = vec(\sum_{i=1}^{m}\sum_{j=1}^{n}{\mF_t(i)}^T\mF_s(j))
% \end{align}

Nevertheless, its efficiency may suffer from the memory overhead of storing high-dimensional features with the quadratic expansion. To solve the problem, we transform the final bilinear feature into a low-rank one by inserting and factorizing an interaction matrix $\mW \in \R^{m\times n}$. It is first inserted to make linear transformation between each temporal-spectral feature pair:
\begin{align}
    \mF_{bilinear} &= {\mF_t}^T\times\mW\times\mF_s \\
    &= \sum_{i=1}^{m}\sum_{j=1}^{n}{\mF_t(i)}^T\mW(i,j)\mF_s(j)
\end{align}
Then, we use $\mW=\mU{\mV}^T$ to factorize the interaction matrix into $\mU \in \R^{m\times l}$ and $\mV^{n\times l}$ ($l<<d$) for obtaining low-rank bilinear feature:
\begin{align}
    \mF_{bilinear} &= {\mF_t}^T\times\mU\times{\mV}^T\times\mF_s \\&= ({\mU}^T\times{\mF_t})\circ({\mV}^T\times\mF_s)
\end{align}
where $\circ$ denotes Hadamard product. BTSF employs the two linear mappings without biases to produce the bilinear representations $\mF_{bilinear} \in \R^{l\times d}$ for a given output dimension $l$. Through this process, the storing memory of na\"ive features of Eq.(\ref{ori-eq}) is reduced largely from $O(d^2)$ to $O(ld)$.

For not forgetting the original temporal and spectral information, the initial temporal feature $\mF_{t}\in \R^{l\times d}$ and spectral feature $\mF_{s}\in \R^{l\times d}$ are both combined with $\mF_{bilinear}$ to enhance the representative capacity. Therefore, the final joint feature representation $\vf\in\R^{l\times d}$ of each augmented time series can expressed as the following:
% \begin{equation}
% \begin{aligned}
% \label{final-eq}
%     \vf &=concat(\vf_{bilinear}^{t}, \vf_{bilinear}^{s}, \vf_{bilinear})\\
%     &= concat({\mF_t}^T\mW_t\mF_{t}, {\mF_s}^T\mW_s\mF_{s}, {\mF_t}^T\mW\mF_{s})\\
%     &= concat({\mU_t}^T{\mF_t}\circ{\mV_t}^T\mF_t, {\mU_s}^T{\mF_s}\circ{\mV_s}^T\mF_s ,  {\mU}^T{\mF_t}\circ{\mV}^T\mF_s)
% \end{aligned}
% \end{equation}
\begin{equation}
\begin{aligned}
\label{final-eq}
    \vf 
    = \sigma({\mW_t}^T\times\mF_{t}+ {\mW_s}^T\times\mF_{s}+ {\mF_t}^T\times\mW\times\mF_s)
\end{aligned}
\end{equation}
where $\mW_t \in \R^{m\times l}$ and $\mV_t \in \R^{m\times l}$ are all linear transformation layers. $\sigma$ is the sigmoid function.  
After vectorizing the feature representations $\vf^{anc}$, $\vf^{pos}$ and $\vf^{neg}$ of a contrastive tuple $(\vx^{anc}, \vx^{pos}, \vx^{neg})$, we build a loss function to minimize and maximize the distance of positive and negative pairs respectively. We represent a multivariate time series as $\mX \in \R^{D \times T}= \{{\vx_j}\}_{j=1}^{D}$, where $D$ is the number of variables and $T$ is the length of time series. Thus, the contrastive loss for a training batch of multivariate time series can be expressed as the following:
% \begin{align}
%     \mathcal{L} = \E_{\mX\sim data}\left[-\E_{(\vx,\vx^{pos})\sim \mX}\left[log(\vf^T\vf^{pos}) \right]+\E_{(\vx,\vx^{neg})\sim\mX}\left[log(\vf^T\vf^{neg})\right]\right]
% \end{align}
% \begin{align}
%     \mathcal{L} = \E_{\mX\sim data}\left[-\E_{(\vx,\vx^{pos})\sim \mX}\left[log(\vf^T\vf^{pos}) \right]+\E_{\vx\sim\mX}\left[log(\E_{\vx^{neg}\sim\mX}\left[\vf^T\vf^{neg}\right])\right]\right]
% \end{align}
% \begin{equation}
% \begin{aligned}
% \label{eq-loss}
%     \mathcal{L} =
%     \E_{\mX\sim \mP_{data}}\left[\E_{(\vx,\vx^{pos})\sim \mX}\left[-log({sim(\vf^{anc},\vf^{pos})}) +\E_{\vx^{neg}\sim\mX}\left[log({sim(\vf^{anc},\vf^{neg})})\right]\right]\right]
% \end{aligned}
% \end{equation}
\begin{equation}
\begin{aligned}
\begin{split}
\label{eq-loss}
    \mathcal{L} =
    &\E_{\mX\sim \mP_{data}}[-log({sim(\vf^{anc},\vf^{pos})/\tau}) +\\ &\E_{\vx^{neg}\sim\mX}\left[log({sim(\vf^{anc},\vf^{neg})/\tau})\right]]
    \end{split}
\end{aligned}
\end{equation}
where $sim(\cdot,\cdot)$ denotes the inner product to measure the distance between two $\ell_2$ normalized feature vectors and $\tau$ is a temperature parameter. Eq.(\ref{eq-loss}) demonstrates that for each multivariate time series, when a time series is chosen for constructing the positive pair, time series of all other variables are the negative samples. For ablation studies of hyperparameters, see Appendix \ref{more ablation}.

\subsection{Effectiveness of the Proposed BTSF}
To prove the efficiency of our devised bilinear fusion, we provide the deduction of gradient flow from the loss function. Since the overall architecture is a directed acyclic graph, the parameters can be trained by back-propagating the gradients of the contrastive loss. The bilinear form simplifies the gradient computations.
Let $\frac{\partial \mathcal{L}}{\partial \vf}$ be the gradient of $\mathcal{L}$ with respect to $\vf$, then for Eq.(\ref{final-eq}) by chain rule of gradients (we omit the sigmoid function for simplicity):
\begin{align}
\begin{split}
\label{gradient-eq1}
    \frac{\partial \mathcal{L}}{\partial \mF_t} = \frac{\partial \mathcal{L}}{\partial \vf}\mW_t + \frac{\partial \mathcal{L}}{\partial \vf}\mW\times\mF_s,\\
    \frac{\partial \mathcal{L}}{\partial \mF_s} = \frac{\partial \mathcal{L}}{\partial \vf}\mW_s + \frac{\partial \mathcal{L}}{\partial \vf}\mW^T\times\mF_t
\end{split}
\end{align}
\begin{align}
\begin{split}
\label{gradient-eq3}
    \frac{\partial \mathcal{L}}{\partial \mW_t}&=\frac{\partial \mathcal{L}}{\partial \vf}\mF_t,\\
    \frac{\partial \mathcal{L}}{\partial \mW_s}&=\frac{\partial \mathcal{L}}{\partial \vf}\mF_s,\\
    \frac{\partial \mathcal{L}}{\partial \mW}&=\frac{\partial \mathcal{L}}{\partial \vf}\mF_t\times{\mF_s}^T
\end{split}
\end{align}
\begin{align}
\begin{split}
\label{gradient-eq2}
    \frac{\partial \mathcal{L}}{\partial \vtheta_t} = \frac{\partial \mathcal{L}}{\partial \vf}\frac{\partial \vf}{\partial \mF_t}\mW_t + \frac{\partial \mathcal{L}}{\partial \vf}\frac{\partial \vf}{\partial \mF_t}\mW\times\mF_s,\\
    \frac{\partial \mathcal{L}}{\partial \vtheta_s} = \frac{\partial \mathcal{L}}{\partial \vf}\frac{\partial \vf}{\partial \mF_s}\mW_s + \frac{\partial \mathcal{L}}{\partial \vf}\frac{\partial \vf}{\partial \mF_s}\mW^T\times\mF_t
\end{split}
\end{align}
From the Eq.(\ref{gradient-eq1}) and Eq.(\ref{gradient-eq2}), , we conclude that the gradient update of parameters $\vtheta_t$ in temporal feature $\mF_t$ is closely related to the spectral feature since $\mF_s$ is treated as a weighted coefficient straightly multiplying the gradient, and vice versa. Additionally, we can know that interaction matrix $\mW$ has a strong connection with cross-domain affinities $\mF_t\times{\mF_s}^T$ from the Eq.(\ref{gradient-eq3}) which leads to a better combination of temporal and spectral features. In hence, it is proved that our BTSF adequately explores and utilizes the underlying spectral and temporal information of time series. 

\section{Experiments}
\label{experiment}
We apply our BTSF on multiple time series datasets in three major practical tasks including classification, anomaly detection and forecasting. We are the first to evaluate on all three tasks. We compare our performances with state-of-the-art approaches CPC, SRL, TS-TCC and TNC. For fair comparisons, we implement these methods by public code with the same encoder architecture and the 
similar computational complexity and parameters, also use the same representation dimensions with BTSF. 
More specific descriptions of tasks definitions, datasets and experiments are in Appendix \ref{dataset disc}. \begin{table*}[ht]
\begin{center}
\begin{small}
\caption{Comparisons of classification results.}
\vskip 0.15in
  \label{cls}
  \center
  \setlength{\tabcolsep}{1.6mm}{
  \begin{tabular}{c|cccccccccc}
  \toprule
\multirow{2}{*}{Methods}  &
\multirow{2}{*}{\space}&
\multicolumn{2}{c}{HAR} &
\multicolumn{1}{c}{\space}& \multicolumn{2}{c}{Sleep-EDF}&
\multicolumn{1}{c}{\space}& \multicolumn{2}{c}{ECG Waveform}&
\\ \cline{3-4} \cline{6-7} \cline{9-10} &&Accuracy & AUPRC &&Accuracy& AUPRC&&Accuracy & AUPRC  \\ \midrule
Supervised&&92.03$\pm$2.48&0.98$\pm$0.00&&83.41$\pm$1.44&0.78$\pm$0.52&&84.81$\pm$0.28&0.67$\pm$0.01\\ \midrule
KNN&&84.85$\pm$0.84&0.75$\pm$0.01&&64.87$\pm$1.73 &0.75$\pm$2.88 &&54.76$\pm$5.46&0.38$\pm$0.06\\
SRL&&63.60$\pm$3.37&0.71$\pm$0.01&& 78.32$\pm$1.45&0.71$\pm$2.83 &&75.51$\pm$1.26&0.47$\pm$0.00\\
CPC&&86.43$\pm$1.41&0.93$\pm$0.01&&82.82$\pm$1.68 & 0.73$\pm$2.15 &&68.64$\pm$0.49&0.42$\pm$0.01\\
TS-TCC&&88.04$\pm$2.46&0.92$\pm$0.02&&83.00$\pm$0.71&0.74$\pm$2.63&&74.81$\pm$1.10&0.53$\pm$0.02\\
TNC&&88.32$\pm$0.12&0.94$\pm$0.01&& 82.97$\pm$0.94&0.76$\pm$1.73 &&77.79$\pm$0.84&0.55$\pm$0.01\\
\textbf{BTSF}&&\textbf{94.63$\pm$0.14}&\textbf{0.99$\pm$0.01}&&\textbf{87.45$\pm$0.54}&\textbf{0.79$\pm$0.74}&&\textbf{85.14$\pm$0.38}&\textbf{0.68$\pm$0.01}\\
\bottomrule
\end{tabular}}
\end{small}
\end{center}
\end{table*}

\begin{table*}[ht]
\caption{Comparisons of multivariate forecasting results.}
\vskip 0.15in
\begin{center}
\begin{small}

  \label{fore}
  \center
  \setlength{\tabcolsep}{1.3mm}{
  \begin{tabular}{cc|ccccccccccccccccccc}
  \toprule
\multirow{2}{*}{Datasets}  &
\multirow{2}{*}{Length} &
\multirow{2}{*}{\space}&
\multicolumn{2}{c}{Supervised} &
\multicolumn{1}{c}{\space}&
\multicolumn{2}{c}{SRL}&
\multicolumn{1}{c}{\space}&
\multicolumn{2}{c}{CPC}&
\multicolumn{1}{c}{\space}&
\multicolumn{2}{c}{TS-TCC}&
\multicolumn{1}{c}{\space}&
\multicolumn{2}{c}{TNC}&
\multicolumn{1}{c}{\space}&
\multicolumn{2}{c}{\textbf{BTSF}}
% \\ \cline{3-4} \cline{6-7} \cline{9-10} \cline{12-13} \cline{15-16}
\\ \cline{4-5} \cline{7-8} \cline{10-11} \cline{13-14} \cline{16-17} \cline{19-20}
&&&MSE & MAE&& MSE & MAE&&MSE & MAE&&MSE & MAE&&MSE & MAE&&MSE & MAE&\\ \midrule
\multirow{5}{*}{ETTh1}&24&&0.577&0.549&&0.698&0.661&&0.687&0.634&&0.653&0.610&&0.632&0.596&&\textbf{0.541}&\textbf{0.519}\\&48&&0.685&0.625&&0.758&0.711&&0.779&0.768&&0.720&0.693&&0.705&0.688&&\textbf{0.613}&\textbf{0.524}\\&168&&0.931&0.752&&1.341&1.178&&1.282&1.083&&1.129&1.044&&1.097&0.993&&\textbf{0.640}&\textbf{0.532}\\&336&&1.128&0.873&&1.578&1.276&&1.641&1.201&&1.492&1.076&&1.454&0.919&&\textbf{0.864}&\textbf{0.689}\\&720&&1.215&0.896&&1.892&1.566&&1.803&1.761&&1.603&1.206&&1.604&1.118&&\textbf{0.993}&\textbf{0.712}\\
\midrule
\multirow{5}{*}{ETTh2}&24&&0.720&0.665&&1.034&0.901&&0.981&0.869&&0.883&0.747&&0.830&0.756&&\textbf{0.663}&\textbf{0.557}\\&48&&1.451&1.001&&1.854&1.542&&1.732&1.440&&1.701&1.378&&1.689&1.311&&\textbf{1.245}&\textbf{0.897}\\&168&&3.389&1.515&&5.062&2.167&&4.591&3.126&&3.956&2.301&&3.792&2.029&&\textbf{2.669}&\textbf{1.393}\\&336&&2.723&1.340&&4.921&3.012&&4.772&3.581&&3.992&2.852&&3.516&2.812&&\textbf{1.954}&\textbf{1.093}\\&720&&3.467&1.473&&5.301&3.207&&5.191&2.781&&4.732&2.345&&4.501&2.410&&\textbf{2.566}&\textbf{1.276}\\\midrule
\multirow{5}{*}{ETTm1}&24&&0.323&0.369&&0.561&0.603&&0.540&0.513&&0.473&0.490&&0.429&0.455&&\textbf{0.302}&\textbf{0.342}\\&48&&0.494&0.503&&0.701&0.697&&0.727&0.706&&0.671&0.665&&0.623&0.602&&\textbf{0.395}&\textbf{0.387}\\&96&&0.678&0.614&&0.901&0.836&&0.851&0.793&&0.803&0.724&&0.749&0.731&&\textbf{0.438}&\textbf{0.399}\\&288&&1.056&0.786&&2.471&1.927&&2.066&1.634&&1.958&1.429&&1.791&1.356&&\textbf{0.675}&\textbf{0.429}\\&672&&1.192&0.926&&2.042&1.803&&1.962&1.797&&1.838&1.601&&1.822&1.692&&\textbf{0.721}&\textbf{0.643}\\\midrule
\multirow{5}{*}{Weather}&24&&0.335&0.381&&0.688&0.701&&0.647&0.652&&0.572&0.603&&0.484&0.513&&\textbf{0.324}&\textbf{0.369}\\&48&&0.395&0.459&&0.751&0.883&&0.720&0.761&&0.647&0.691&&0.608&0.626&&\textbf{0.366}&\textbf{0.427}\\&168&&0.608&0.567&&1.204&1.032&&1.351&1.067&&1.117&0.962&&1.081&0.970&&\textbf{0.543}&\textbf{0.477}\\&336&&0.702&0.620&&2.164&1.982&&2.019&1.832&&1.783&1.370&&1.654&1.290&&\textbf{0.568}&\textbf{0.487}\\&720&&0.831&0.731&&2.281&1.994&&2.109&1.861&&1.850&1.566&&1.401&1.193&&\textbf{0.601}&\textbf{0.522}\\
% \\ \cline{4-5} \cline{7-8} \cline{10-11} \cline{13-14} \cline{16-17} \cline{19-20}
% &&&MSE & MAE&& MSE & MAE&&MSE & MAE&&MSE & MAE&&MSE & MAE&&MSE & MAE&\\ \midrule
% \multirow{2}{*}{ETTh1}&48&&0.685&0.625&&0.758&0.711&&0.779&0.768&&0.720&0.693&&0.705&0.688&&\textbf{0.613}&\textbf{0.524}\\&168&&0.931&0.752&&1.341&1.178&&1.282&1.083&&1.129&1.044&&1.097&0.993&&\textbf{0.640}&\textbf{0.532}\\&720&&1.215&0.896&&1.892&1.566&&1.803&1.761&&1.603&1.206&&1.604&1.118&&\textbf{0.993}&\textbf{0.712}\\
% \midrule
% \multirow{2}{*}{ETTh2}&48&&1.451&1.001&&1.854&1.542&&1.732&1.440&&1.701&1.378&&1.689&1.311&&\textbf{0.544}&\textbf{0.527}\\&168&&3.389&1.515&&5.062&2.167&&4.591&3.126&&3.956&2.301&&3.792&2.029&&\textbf{1.669}&\textbf{0.875}\\&720&&3.467&1.473&&5.301&3.207&&5.191&2.781&&4.732&2.345&&4.501&2.410&&\textbf{2.566}&\textbf{1.276}\\\midrule
% \multirow{2}{*}{ETTm1}&48&&0.494&0.503&&0.701&0.697&&0.727&0.706&&0.671&0.665&&0.623&0.602&&\textbf{0.395}&\textbf{0.387}\\&96&&0.678&0.614&&0.901&0.836&&0.851&0.793&&0.803&0.724&&0.749&0.731&&\textbf{0.438}&\textbf{0.399}\\&672&&1.192&0.926&&2.042&1.803&&1.962&1.797&&1.838&1.601&&1.822&1.692&&\textbf{0.721}&\textbf{0.643}\\\midrule
% \multirow{2}{*}{Weather}&48&&0.395&0.459&&0.751&0.883&&0.720&0.761&&0.647&0.691&&0.608&0.626&&\textbf{0.366}&\textbf{0.427}\\&168&&0.608&0.567&&1.204&1.032&&1.351&1.067&&1.117&0.962&&1.081&0.970&&\textbf{0.543}&\textbf{0.477}\\&720&&0.831&0.731&&2.281&1.994&&2.109&1.861&&1.850&1.566&&1.401&1.193&&\textbf{0.601}&\textbf{0.522}\\

\bottomrule
\end{tabular}}

\end{small}
\end{center}
\end{table*}

\begin{table*}[ht]
\caption{Comparisons of multivariate anomaly detection.}
\vskip 0.15in
  \label{anomaly}
  \center
\begin{small}

  \begin{tabular}{cc|ccccccc}
  \toprule
{Datasets}  &
{Metric} &\space&Supervised  & SRL & CPC & TS-TCC& TNC & \textbf{BTSF}\\ \midrule

\multirow{1}{*}{SAaT}&F1&&0.901&0.710&0.738&0.775&0.799&\textbf{0.944}\\\midrule
\multirow{1}{*}{WADI}&F1&&0.649&0.340&0.382&0.427&0.440&\textbf{0.685}\\\midrule
\multirow{1}{*}{SMD}&F1&&0.958&0.768&0.732&0.794&0.817&\textbf{0.972}\\\midrule
\multirow{1}{*}{SMAP}&F1&&0.842&0.598&0.620&0.679&0.693&\textbf{0.863}\\\midrule
\multirow{1}{*}{MSL}&F1&&0.945&0.788&0.813&0.795&0.833&\textbf{0.957}\\
\bottomrule
\end{tabular}
% }
\end{small}
\end{table*}

% \begin{table*}
% \caption{Comparisons of classification results on all UCR and UEA datasets.}
%   \label{cls other2}
%   \center
%   \setlength{\tabcolsep}{1.0mm}{
%   \begin{tabular}{c|cccccccccc}
%   \toprule
% \multirow{2}{*}{Methods}  &
% \multirow{2}{*}{\space}&
% \multicolumn{2}{c}{UCR datasets} &
% \multicolumn{1}{c}{\space}& \multicolumn{2}{c}{UEA datasets}

% \\ \cline{3-4} \cline{6-7} &&Average Accuracy & Average Rank &&Average Accuracy & Average Rank\\ \midrule
% Supervised&&89.67&1.67&&82.04&1.54\\ \midrule
% KNN&&73.81&4.68&&67.81&2.77\\
% SRL&&81.24&3.27&&68.30&2.37\\
% CPC&&80.57&3.54&&65.84&2.59\\
% TS-TCC&&82.75&2.68&&69.43&2.11\\
% TNC&&79.95&2.27&&70.58&2.25\\
% \textbf{BTSF}&&\textbf{92.11}&\textbf{1.33}&&\textbf{86.72}&\textbf{1.26}\\
% \bottomrule
% \end{tabular}}
% \end{table*}
\paragraph{Time-Series Classification.}

We evaluate our learned representation on downstream classification tasks for time series on widely-used time series classification datasets
\citep{anguita2013public,goldberger2000physiobank,andrzejak2001indications,moody1983new}.
For fair comparisons, we further train a linear classifier on top of the learned representations to evaluate how well the representations can be used to classify hidden states, following \citet{tonekaboni2021unsupervised}. Beyond aforementioned methods, we also implement a K-nearest neighbor classifier equipped with DTW \citep{chen2013dtw} metric and a supervised model which is trained with the same encoder and classifier with those of our unsupervised model. In the training stage, we keep the original train/test splits of datasets and use the training set to train all the models. We apply two metrics for evaluation, the prediction accuracy and the area under the precision-recall curve (AUPRC). Table \ref{cls} demonstrates our superior performance over existing methods in all datasets and our BTSF surpasses the supervised method, which shows that BTSF adequately leverages the temporal and spectral information in time series for representation learning. In addition, the pair-wise temporal-spectral fusion provides more fine-grained information for discriminativity.

\paragraph{Time-Series Forecasting.}
We evaluate our algorithm with other methods on time series forecasting task in both short-term and long-term settings, following \citet{zhou2021informer}. A decoder is added on top of learned representations to make predictive outputs. Specifically, we train a linear regression model with L2 norm penalty and use informer \citep{zhou2021informer} as our supervised comparison method. We use two metrics to evaluate the forecasting performance, Mean Square Error (MSE) and Mean Absolute Error (MAE). Table \ref{fore} demonstrates that our BTSF has the least forecasting error of different prediction lengths (short/long) across the datasets. In addition, BTSF outperforms existing methods including supervised one in a large margin especially for long time series prediction. It is noted that BTSF gets a better performance when the length of datasets increases due to the better use of global context, which makes BTSF fully capture the long-term dependencies in long time series. More visualization results of time series forecasting are in Appendix \ref{app-fore}, Fig.\ref{fig:foree} and Fig.\ref{fig:foree2}.

\paragraph{Time-Series Anomaly Detection.}
To the best of our knowledge, we are the first to evaluate on anomaly detection \citep{su2019robust,hundman2018detecting,goh2016dataset,mathur2016swat,braei2020anomaly}. The results of this task assessment reflect how well the model capture the temporal trends and how sensitive to the outlier the model is for time series. We add a decoder on top of representations learned by models and reconstruct the input time series and follow the evaluation settings of \citet{audibert2020usad}. 
For each input data point $\vx_t$ and reconstructed one $\hat{\vx}_{t}$, if $|\hat{\vx}_{t}-\vx_t|>\tau$ ($\tau$ is a predefined threshold), $\vx_t$ is an outlier. Precision (P), Recall (R), and F1 score (F1) were used to evaluate
anomaly detection performance and we just list the results of F1 metric here (see Appendix \ref{dataset disc} for more results of P and R metrics). 
Table \ref{anomaly} illustrates that BTSF achieves new SOTA across all datasets and especially surpasses the supervised results by a large margin. It conveys that BTSF is more sensitive to the outliers in time series since it captures long-term dynamics and expresses the fine-grained information through iterative bilinear fusion. 

\section{Analysis}
\begin{figure}[ht]
\centering
    \includegraphics[width=6.0cm]{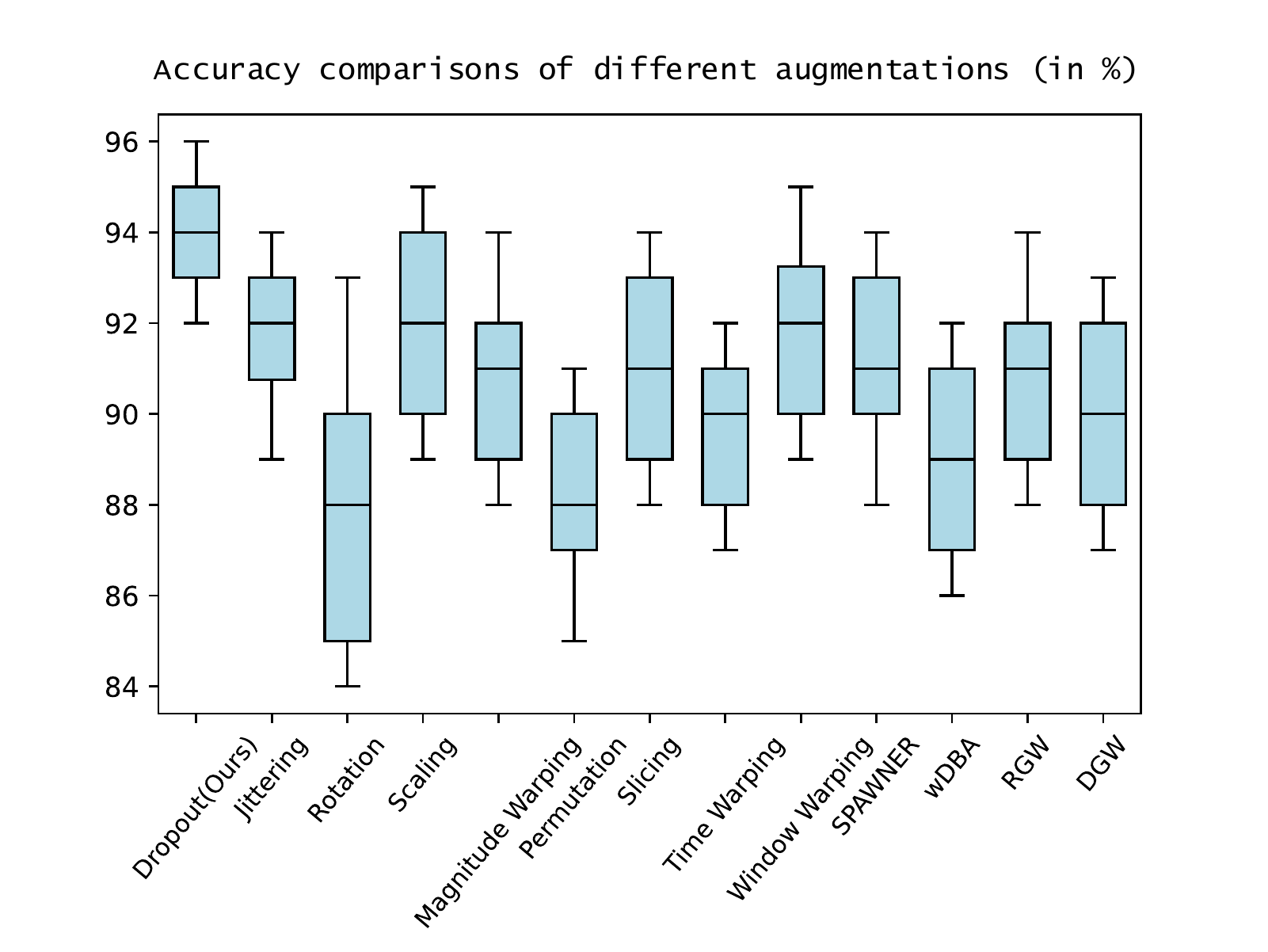}
\caption{Classification accuracies and variances of different augmentations on HAR dataset.}
\label{fig:aug}
\end{figure}

\paragraph{Comparisons about Time-Series Augmentation Methods.}To further prove the effectiveness of our instance-level augmentation (dropout), we compare our method with 12 other augmentation policies as mentioned in \citet{iwana2021empirical}: Jittering, Rotation, Scaling, Magnitude Warping,  Permutation, Slicing, Time Warping, Window Warping, SPAWNER \citep{kamycki2020data}, 
Weighted DTW Barycentric Averaging (wDBA) \citep{forestier2017generating}, 
Random Guided Warping (RGW) \citep{iwana2021time} and
Discriminative Guided Warping (DGW) \citep{iwana2021time}. 
The classification accuracy comparisons of different augmentations on HAR datasets are illustrated in Figure \ref{fig:aug}. It is noted that proposed instance-level 
augmentation (dropout) has a best performance in both average accuracy and variance, which demonstrates dropout is more accurate and more stable for unsupervised representation learning in time series.
\paragraph{Impact of Iterative Bilinear Fusion}
To investigate the impact of iterative bilinear fusion in BTSF, we follow the experiment as illustrated in Section \ref{intro1}. We apply the learned representations of models to the classification task and make statistics about false predictions by only using temporal or spectral feature respectively. Specifically, we use the feature out of $\operatorname{S2T}$ and $\operatorname{T2S}$ module as temporal and spectral feature respectively. From Table \ref{state}, we find that after adding iterative bilinear fusion, BTSF not only gets a large promotion in accuracy but also achieves a good alignment between temporal and spectral domain with a overlap percentage of \textbf{96.60\%}, much higher than existing works (around \textbf{30\%}). Therefore, our designed iterative bilinear fusion make an effective interaction between two domains and it is vital for final prediction accuracy.
More ablation studies about BTSF are in Appendix \ref{more ablation}.
\begin{table*}[ht]
  \caption{Statistics about false predictions of all test samples on HAR dataset }
  \vskip 0.1in
  \label{state}
  \center
  \setlength{\tabcolsep}{2.0mm}{
  \begin{tabular}{cccccc}
    \toprule

     & Only Temporal &Only Spectral &Overlap (\% by Temporal, \% by Spectral)\\\midrule
    SRL&1073&1174&349 (32.53\%, 29.73\%)\\
    CPC&401&448&106 (26.43\%, 23.66\%)\\
    TS-TCC&354 &383&107 (30.23\%, 27.94\%) \\
    TNC  & 346&376&115 (33.24\%, 30.59\%)\\
    \textbf{BTSF}&\textbf{159}&\textbf{163}&\textbf{152} (\textbf{96.60\%}, \textbf{93.25\%}) \\ \bottomrule
 \end{tabular}
 }
 \vskip 0.1in
 \end{table*}

\begin{figure}[ht]
\vskip 0.1in
\centering
    \includegraphics[width=8cm]{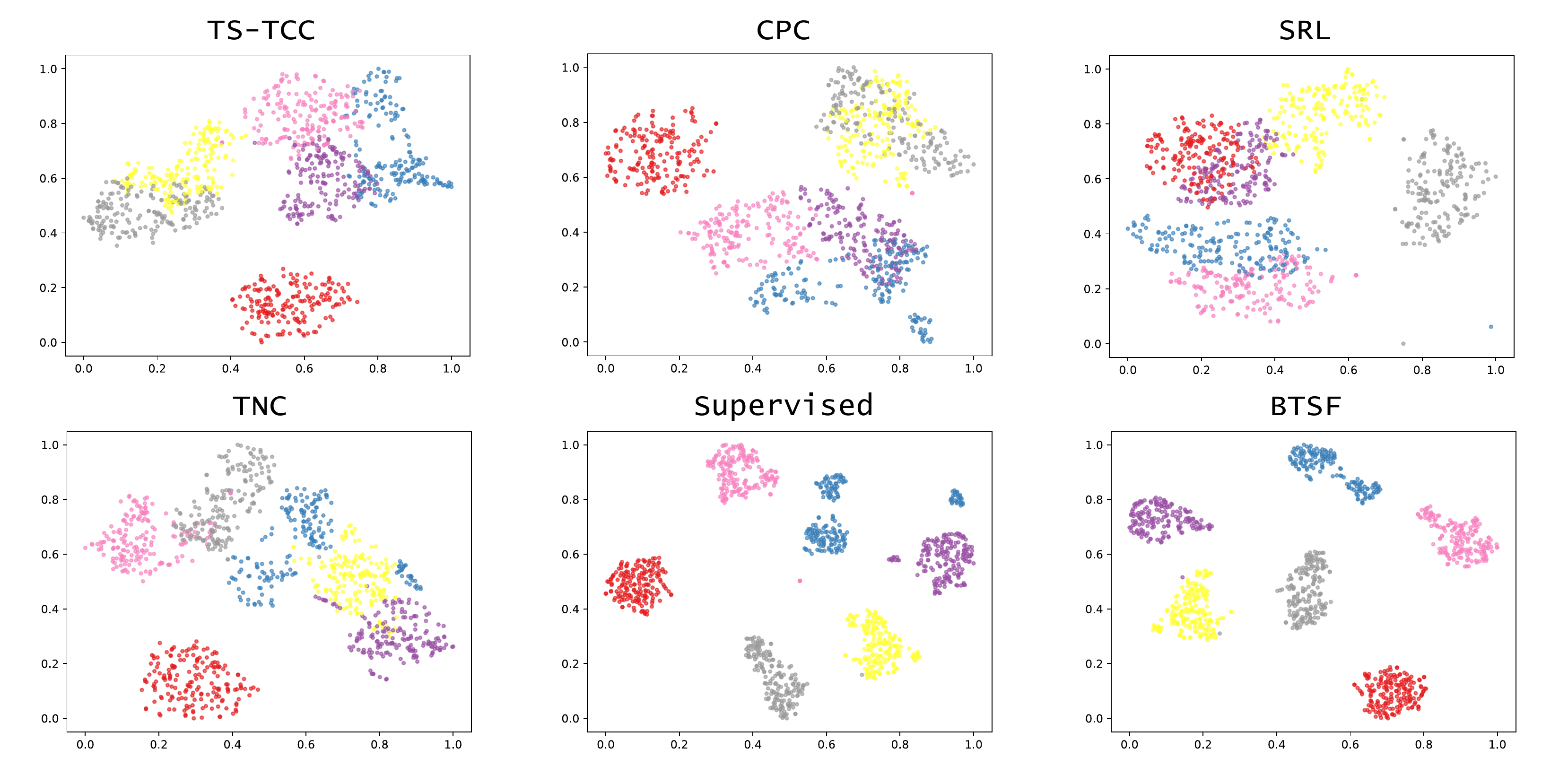}
\caption{T-SNE visualization of signal representations for HAR dataset.}
\label{fig:tsne}
\end{figure}

\begin{figure}[h]
\centering
    \includegraphics[width=8cm]{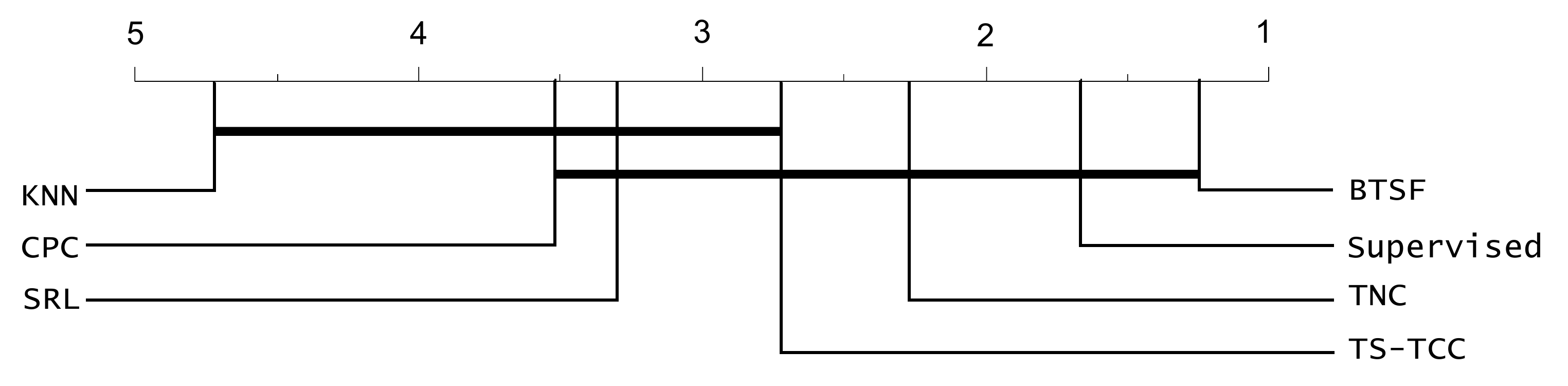}
\caption{Critical difference diagram showing pairwise statistical difference comparison of BTSF and previous methods on the UCR archive.}
\label{fig:critical}
\end{figure}

\paragraph{Visualization.}
\label{tsne-vis}
To make assessments about the clusterability of learned representations in the encoding space, we visualize the feature distribution by using t-SNE \citep{van2008visualizing}. It is noted that if the information of the latent state is properly learned and encoded by the model, the representations from the same underlying state should cluster together. Figure \ref{fig:tsne} shows the comparisons about representations distribution of different models. It demonstrates that the representations learned by proposed BTSF from the same hidden state are better than the other approaches. The visualization results further prove the superior representation ability of our model. In Addition, we have evaluated on the all univariate time series datasets: the UCR archive. The corresponding critical difference diagram is shown in Figure \ref{fig:critical}. The BTSF significantly outperforms the other approaches with an average rank of almost 1.3.

\begin{figure}[h]
\centering
    \includegraphics[width=8cm]{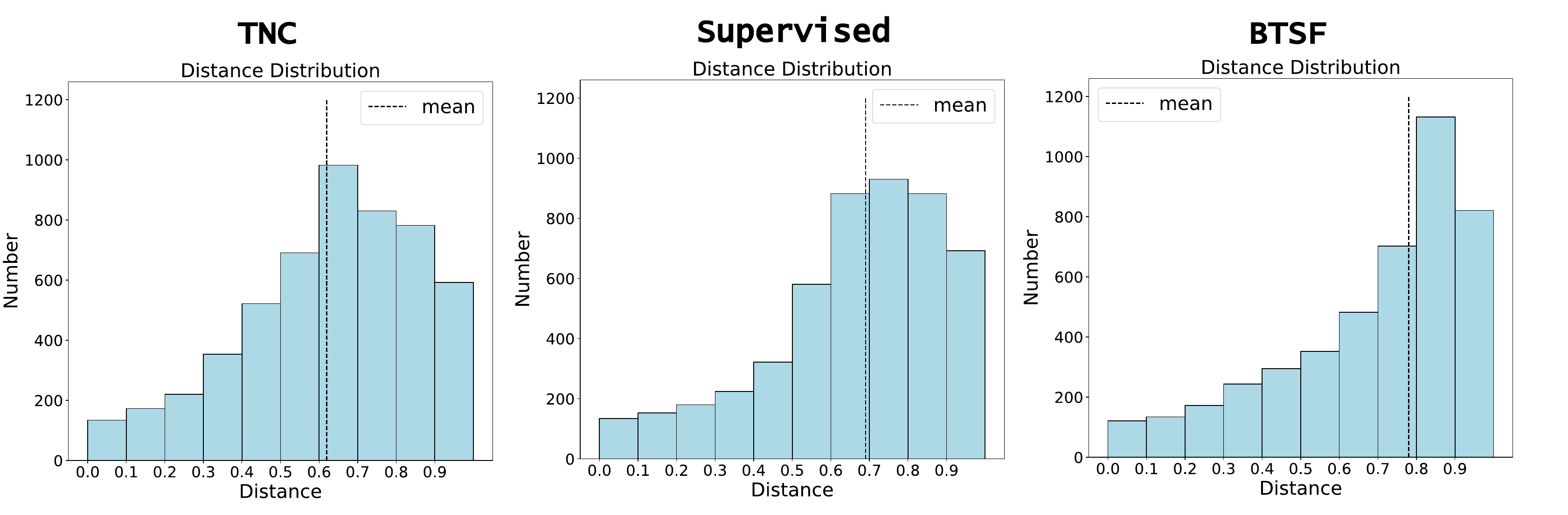}
\vskip 0.1in
\caption{Distance distribution of positive pairs for assessing alignment. Our BTSF is well aligned.}
\label{fig:align}
\end{figure}

\begin{figure}[h]

\centering
    \includegraphics[width=8cm]{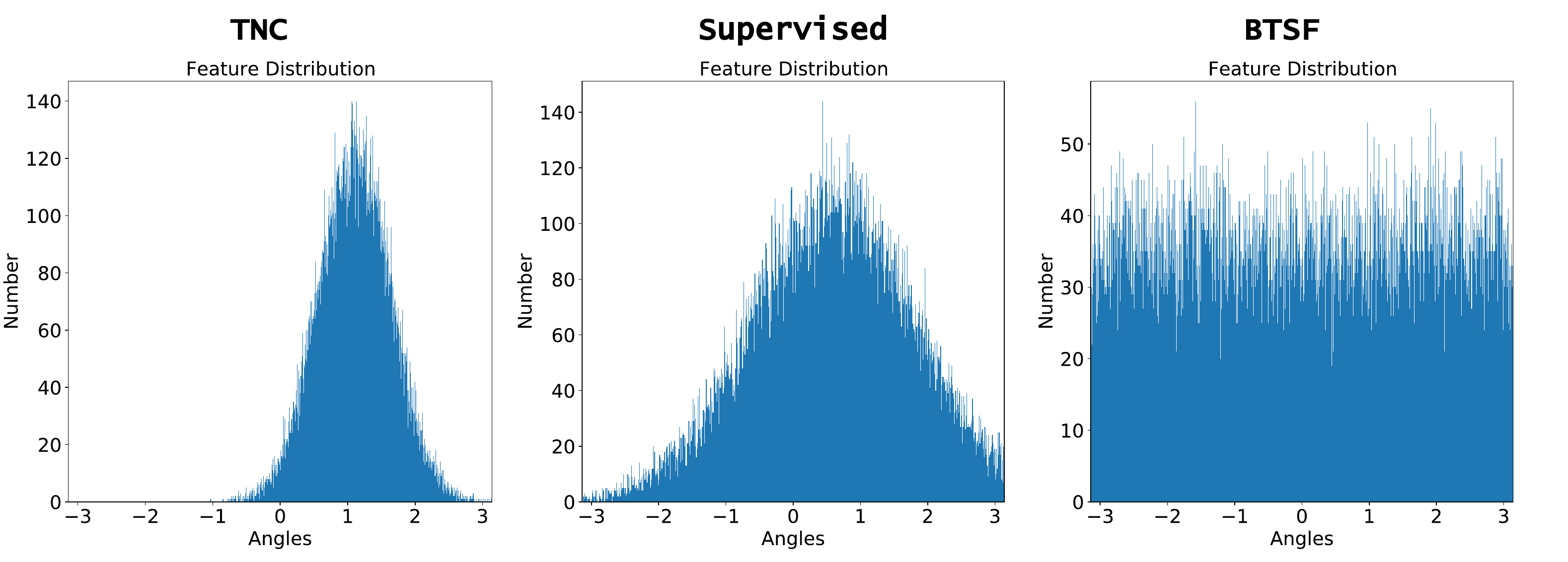}
\vskip 0.1in
\caption{Feature distribution of samples in different classes on the normalized surface area for assessing uniformity. Features extracted by BTSF are evenly distributed.}
\label{fig:uniform}
\end{figure}

\paragraph{Alignment and Uniformity.}
\label{align and uniform}
To make a comprehensive assessment of the representations, we evaluate the two properties of learned representations, \textit{alignment} and \textit{uniformity} \citep{wang2020understanding}. \textit{Alignment} is used to measure the similarities of features between similar samples, which means features of a positive pair should be invariant to the noise. \textit{Uniformity} assumes that a well-learned feature distribution should preserve maximal information as much as possible. 
It makes sense that well-generalized feature representations not only minimize the intra-similarities of positive pairs and enlarge the inter-distances of negative pairs but also keep the feature distributed uniformly to retain enough information. Therefore we follow \citet{wang2020understanding} to make the assessments. Figure \ref{fig:align} and Figure \ref{fig:uniform} show the results of alignment and uniformity respectively. Compared with previous SOTA TNC and supervised results, our BTSF gets the highest mean value about feature distance of positive pairs, which means that BTSF achieves the best alignment. Additionally, the feature extracted BTSF is evenly distributed in the encoding space which preserves maximal information of the data, much better than TNC and supervised models.

\section{Conclusion}
In this paper, we propose Bilinear Temporal-Spectral Fusion (BTSF) for unsupervised time series representation learning. We revisit existing segment-level contrastive learning methods and conclude that they all fail to leverage global contextual information due to the segment-level augmentation (time slicing) and are unable to use temporal-spectral relations for enhancing representation learning. First, we utilize instance-level augmentation which use the entire time series as input and apply dropout to generate different views for training. Second, we devise \textit{iterative bilinear temporal-spectral fusion} and refine the feature representation in a fuse-and-squeeze manner for time series. The extensive experiments on classification, forecasting and anomaly detection downstream tasks have been conducted and the results demonstrates the superior performance of our BTSF. BTSF surpasses existing unsupervised learning models for time series in a large margin including the supervised model.  

\section*{Acknowledgement}
This work was supported by the National Natural Science Foundation of China (No.62102008).

\nocite{langley00}

\bibliography{example_paper}
\bibliographystyle{icml2022}

%%%%%%%%%%%%%%%%%%%%%%%%%%%%%%%%%%%%%%%%%%%%%%%%%%%%%%%%%%%%%%%%%%%%%%%%%%%%%%%
%%%%%%%%%%%%%%%%%%%%%%%%%%%%%%%%%%%%%%%%%%%%%%%%%%%%%%%%%%%%%%%%%%%%%%%%%%%%%%%
% APPENDIX
%%%%%%%%%%%%%%%%%%%%%%%%%%%%%%%%%%%%%%%%%%%%%%%%%%%%%%%%%%%%%%%%%%%%%%%%%%%%%%%
%%%%%%%%%%%%%%%%%%%%%%%%%%%%%%%%%%%%%%%%%%%%%%%%%%%%%%%%%%%%%%%%%%%%%%%%%%%%%%%
\newpage
\appendix
\onecolumn

\section{More ablation studies}
\label{more ablation}
To quantify the promotion of each module in BTSF, we make a specific ablation study where all experiments are conducted on HAR dataset and results are in Table \ref{dropout}. We use TNC as a baseline which applies time slicing as augmentation with accuracy of 88.3\%. We could find that our instance-level augmentation (dropout) is better than segment-level augmentation (slicing) and layer-wise dropout (adding dropout in internal layers) has a promotion by 1.5\% compared with slicing. However, we do not apply layer-wise dropout in aforementioned experiments for fair comparisons otherwise our BTSF will have better performance. Besides, incorporating spectral feature with temporal feature by using summation or concatenation will also improve the results, which illustrates the necessity of cross-domain interaction. 
The accuracy is obviously promoted by 2\%$\sim$3\% when involving temporal and spectral information with bilinear fusion, and iterative operation will further improve the performance by enhancing and refining the temporal-spectral interaction. 
In conclusion, instance-level augmentation (dropout) and iterative bilinear fusion are two main modules of BTSF which largely improve the generalization ability of unsupervised learned representations with accuracy of 94.6\%, an improvement of 6.3\% to baseline. 
\begin{table}[ht]
  \caption{Ablation experiments of BTSF.}
  \vskip 0.1in
  \label{dropout}
  \center
  \setlength{\tabcolsep}{1.2mm}{
  \begin{tabular}{cccccc}
    \toprule
    Accuracy & Temporal & Spectral & Sum/Concat & Bilinear & Iterative Bilinear \\ \midrule
    Slicing &88.3 &86.7&88.7&90.7&91.5\\
    Dropout &89.4&88.4&89.8&92.4&\textbf{94.6}\\
    Layer-Wise Dropout &89.8&89.1&90.4&93.1&\textbf{95.4}\\\bottomrule
 \end{tabular}
 }
 \end{table}

\paragraph{Studies of hyperparameters}
In the proposed BTSF, there are some hyperparameters needed to be carefully set, the dropout rate, temperature number $\tau$ and the loops number of iterative bilinear fusion. Table \ref{dropout ablation} illustrates that when the rate is set to 0.1, BTSF acquires the best performance since setting too high value would lose the original properties of time series and setting too low value would bring about representation collapse. Table \ref{tau ablation} demonstrates that when $\tau$ is set to 0.05 , BTSF has the best performance. It is reasonable that proper value of $\tau$ would promote the optimization of training process and make representations more discriminative with the adjustment. We also run the experiments of loops number of iterative bilinear fusion, and we conclude that our iterative bilinear fusion is effective and its performance converges after just three loops.

\begin{table}[ht]
  \caption{Ablation experiments of dropout rate}
  \vskip 0.1in
  \label{dropout ablation}
  \center
  \setlength{\tabcolsep}{1.4mm}{
  \begin{tabular}{cccccccccc}
    \hline
    dropout rate &  p=0.01 & p=0.05 & p=0.1 &p=0.15&  p=0.2 & p=0.3 \\ \hline
    HAR  &90.29&92.78&\textbf{94.63}&93.36&91.21&88.07\\\hline
    Sleep-EDF  &82.76&85.34&\textbf{87.45}&86.01&83.44&80.92\\\hline
 \end{tabular}
 }
 \end{table}
 
\begin{table}[ht]
  \caption{Ablation experiments on temperature number $\tau$.}
  \vskip 0.1in
  \label{tau ablation}
  \center
  \setlength{\tabcolsep}{1.8mm}{
  \begin{tabular}{cccccccccc}
    \hline
    $\tau$  &0.001& 0.01& 0.05& 0.1& 1 \\ \hline
    HAR  &90.04&92.91&\textbf{94.63}&93.04&91.85\\\hline
    Sleep-EDF&82.69&84.82&\textbf{87.45}&85.11&83.28\\\hline
 \end{tabular}
 }
 \end{table}

\section{Datasets descriptions and more experiments}
\label{dataset disc}
In all experiments, we use Pytorch 1.8.1 \citep{paszke2017automatic} and train all the models on a GeForce RTX 2080 Ti GPU with CUDA 10.2. We apply an Adam optimizer \citep{kingma2017adam} with a learning rate of 3e-4, weight decay of 1e-4 and batch size is set to 256. 
In this part, we would introduce all the datasets used in our experiments which involve three kinds of downstream tasks, time series classification, forecasting and anomaly detection. 
The definitions of downstream tasks are detailed in the following:
\begin{itemize}
    \item \textbf{Time Series Classification}: Given the univariate time series $\{x_{1},x_{2},\dots,x_{T}\}$ or multivariate time series $\{\vx_{1},\vx_{2},\dots,\vx_{D}\}$ as input, time series classification is to classify the input consisting of real-valued observations to a certain class.
    \item \textbf{Time Series Forecasting}: Given the past univariate observations $\{x_{t-T_1+1},\dots,x_{t}\}$ or multivariate ones $\{\vx_{t-T_1+1},\dots,\vx_{t}\}$ as input, time series forecasting aims to predict the future data points $\{x_{t+1},x_{t+2},\dots,x_{t+T_2}\}$ or $\{\vx_{t+1},\vx_{t+2},\dots,\vx_{t+T_2}\}$ based on the input.
    \item \textbf{Time Series Anomaly Detection}: Given the univariate time series $\{x_{1},x_{2},\dots,x_{T}\}$ or multivariate time series $\{\vx_{1},\vx_{2},\dots,\vx_{D}\}$ as input, time series anomaly detection is to find out which point ($\hat{x_i}$ or $\hat{\vx_i}$) or subsequence ($\{\hat{x_{1}},\hat{x_{2}},\dots,\hat{x_{T}}\}$ or $\{\hat{\vx_{1}},\hat{\vx_{2}},\dots,\hat{\vx_{T}}\}$) of the input behaves unusually when compared either to the other values in the time series (global outlier) or to its neighboring points (local outlier).
\end{itemize}

\paragraph{Data Preprocessing} Following \citet{franceschi2019unsupervised,zhou2021informer}, for univariate time series classification task, we normalize datasets using z-score so that the set of observations for each dataset has zero mean and unit variance. For multivariate time series classification task, each variable is normalized independently using z-score. For forecasting tasks, all reported metrics are calculated based on the normalized time series.

\subsection{Classification}
In the time series classification task, we choose several popular benchmarks which are widely used in previous works. They are Human Activity Recognition (HAR) \citep{anguita2013public}, Sleep Stage Classification (Sleep-EDF) \citep{goldberger2000physiobank}, Epilepsy Seizure Prediction \citep{andrzejak2001indications}, ECG Waveform \citep{moody1983new}. The detailed introduction to these datasets are as follows: 
\paragraph{Human Activity Recognition}
HAR dataset contains 30 individual subjects which provide six activities for each subject. These six activities are walking, walking upstairs, downstairs, standing, sitting, and lying down. The data of HAR is collected by sensors with a sampling rate of 50 HZ and the collected signals record the continuous activity of every subject.
\paragraph{Sleep Stage Classification}
The dataset is designed for EEG signal classification task where each signal belongs to one of five categories: Wake (W), Non-rapid eye movement (N1, N2, N3) and Rapid Eye Movement (REM). And the Sleep-EDF dataset collects the PSG for the whole night, and we just used a single EEG channel, following previous works \citep{eldele2021attention}.
\paragraph{Epilepsy Seizure Prediction}
The Epileptic Seizure Prediction dataset contains EEG signals which are collected from 500 subjects. The brain activity for each subject was recorded for 23.6 seconds. Additionally, the original classes of the dataset are five, and we preprocess the dataset for classification task like \citet{eldele2021time}.
\paragraph{ECG Waveform}
The ECG Waveform is a real-world clinical dataset, it includes 25 long-term Electrocardiogram (ECG) recordings (10 hours in duration) of human subjects with atrial fibrillation. Besides, it contains two ECG signals with a sampling rate of 250HZ.

% \paragraph{Fault Diagnosis}
% The Fault Diagnosis dataset is also a real-world dataset. It was collected under four different working conditions. Each condition can be viewed as a separate domain since there exists a large difference between these conditions. Each domain has three classes, two fault classes and a healthy class.

% \paragraph{UCR and UEA}
% The UCR and UEA are widely used public datasets for time series analysis. The UCR archive consists of univariate datasets while UEA archive contains multivariate datasets, which cover multiple scenes in real world.
\begin{table}[ht]
\caption{More comparisons of classification results about BTSF and previous work, results of TST \citep{zerveas2021transformer}, Rocket \citep{dempster2020rocket} and Supervised \citep{zerveas2021transformer} are quoted from TST for fair comparisons.}
\vskip 0.1in
  \label{cls other2}
  \center
  \setlength{\tabcolsep}{1.8mm}{
  \begin{tabular}{c|cccccccccc}
  \toprule
Methods&TST&Rocket&Supervised&\textbf{BTSF}\\\midrule
EthanolConcentration&32.6&45.2&33.7&\textbf{49.4}\\
FaceDetection &68.9&64.7&68.1&\textbf{73.0}\\
Handwriting &35.9 &58.8&30.5&\textbf{62.3}\\
Heartbeat &77.6&75.6&77.6 &\textbf{84.7}\\
JapaneseVowels &99.7&96.2&99.4&\textbf{99.8}\\
InsectWingBeat &68.7& -&68.4&\textbf{78.3}\\
PEMS-SF &89.6&75.1&91.9 &\textbf{95.7}\\
SelfRegulationSCP1& 92.2&90.8&92.5&\textbf{96.5} \\
SelfRegulationSCP2 &60.4 &53.3&58.9&\textbf{64.9}\\
SpokenArabicDigits &99.8&71.2&99.3&\textbf{99.8}\\
UWaveGestureLibrary& 91.3&94.4&90.3&\textbf{97.1}\\\midrule
\textbf{Avg Accuracy}&74.8&72.5&74.2&\textbf{82.0}\\\midrule
\textbf{Avg Rank}&1.7&2.3&1.7&\textbf{1.2}\\
\bottomrule
\end{tabular}}
\end{table}
\begin{table}
\caption{More comparisons of classification results of ESP dataset.}
\vskip 0.1in
  \label{cls other1}
  \center
  \setlength{\tabcolsep}{1.4mm}{
  \begin{tabular}{c|cccccccccc}
  \toprule
\multirow{2}{*}{Methods}  &
\multirow{2}{*}{\space}&
\multicolumn{2}{c}{Epilepsy Seizure Prediction}

\\ \cline{3-4}  &&Accuracy & AUPRC \\ \midrule
Supervised&&96.32$\pm$0.38&0.97$\pm$0.65\\ \midrule
KNN&&87.96$\pm$1.32&0.89$\pm$1.04\\
SRL&&94.65$\pm$0.97&0.95$\pm$0.86\\
CPC&&96.61$\pm$0.43&0.97$\pm$0.69\\
TS-TCC&&97.23$\pm$0.10&0.98$\pm$0.21\\
TNC&&96.15$\pm$0.33&0.96$\pm$0.45\\
\textbf{BTSF}&&\textbf{99.01$\pm$0.12}&\textbf{0.99$\pm$0.06}\\
\bottomrule
\end{tabular}}
\end{table}
Table \ref{cls other2} shows the comparison results between BTSF with recent works following their evaluation protocols. The results show that BTSF significantly outperforms them in a large margin.
Table \ref{cls other1}  shows the classification results of Epileptic Seizure Prediction datasets. From the illustrated results, we conclude that our BTSF gets the best performance and exceeds other methods by a large margin in univariate and multivariate time series classification tasks. 

\subsection{Forecasting}
\label{app-fore}
\begin{figure}[ht]
\centering
    \includegraphics[width=10cm]{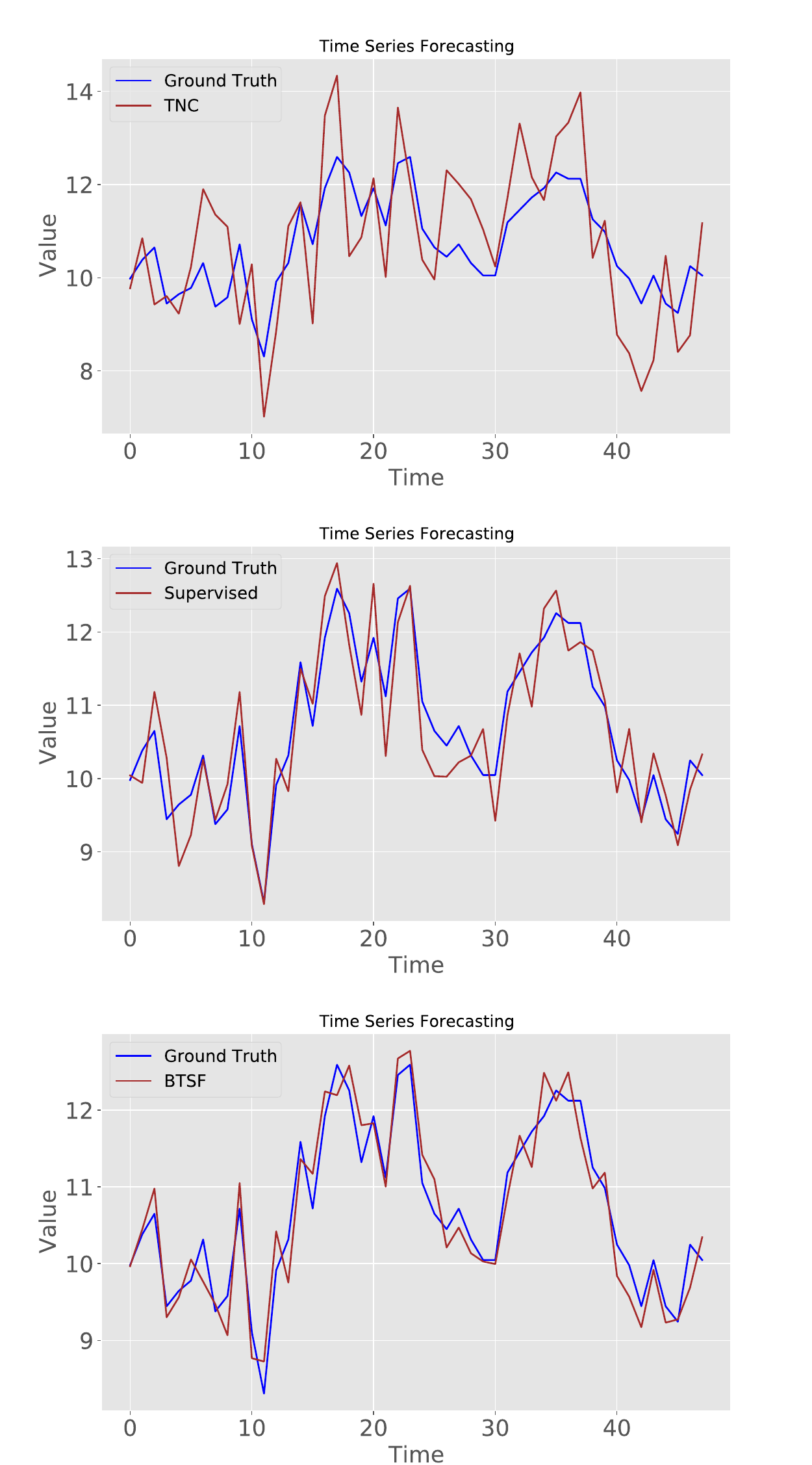}
\caption{Visualizing forecasting results of length 48 on ETT dataset.}
\label{fig:foree}
\end{figure}

\begin{figure}[ht]
\centering
    \includegraphics[width=11cm]{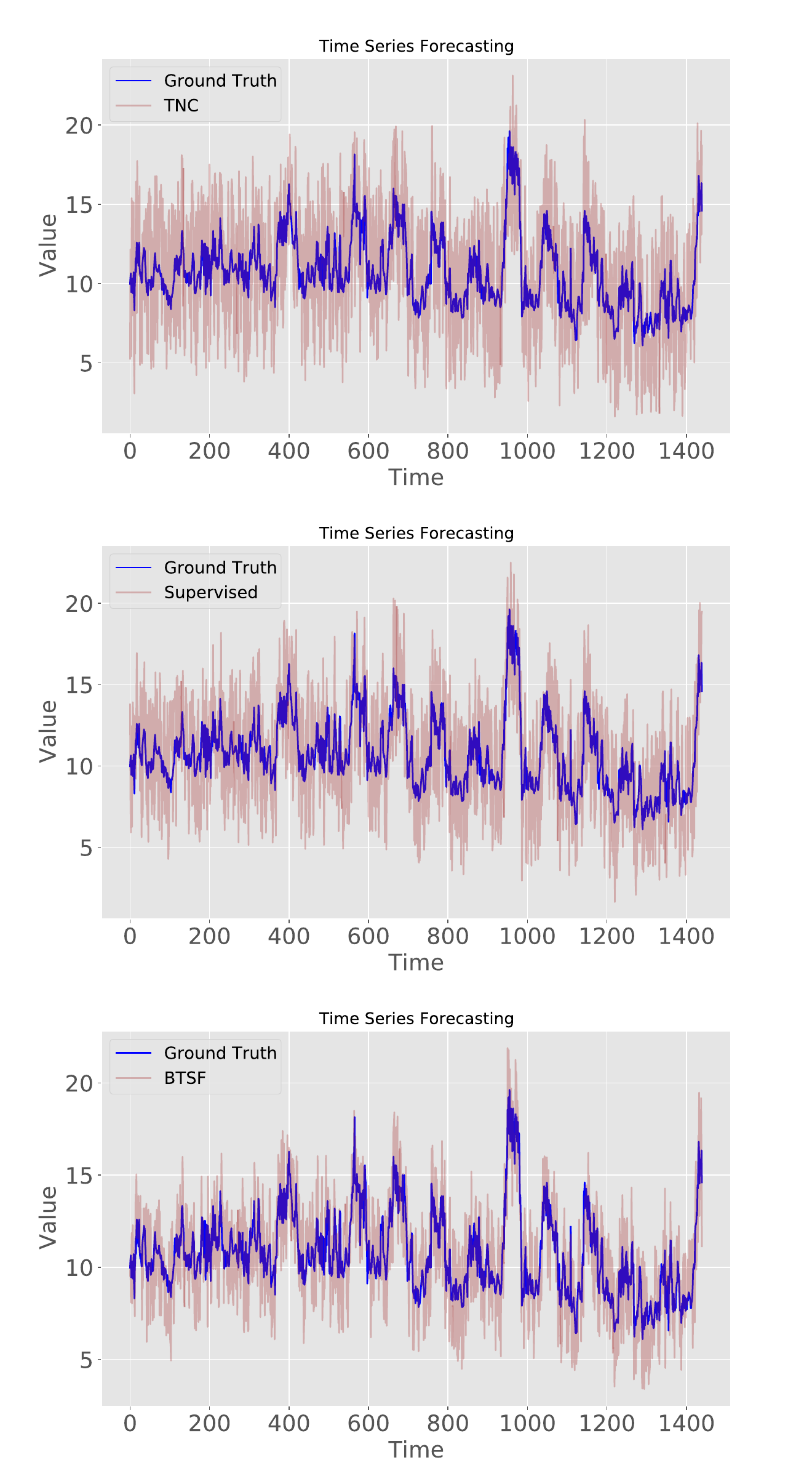}
\caption{Visualizing long-term forecasting results of length 1440 on ETT dataset.}
\label{fig:foree2}
\end{figure}
In Section \ref{experiment}, we conduct experiments on four datasets about time series forecasting, including two collected real-world datasets for long sequence time-series forecasting (LSTF) problem and one public benchmark dataset as in \citet{zhou2021informer}. The detailed introduction to these datasets are as follows:
\paragraph{Electricity Transformer Temperature (ETT)} 
The ETT is a crucial indicator in the electric power long-term deployment. The 2-year data was collected from two separated counties in China, which was first used to investigate the granularity on the LSTF problem with each data point containing the target value ”oil temperature” and six power load features. {ETTh1 , ETTh2} and ETTm1 represent for 1-hour-level and 15-minute-level respectively.

\paragraph{Weather}
This dataset contains local climatological data for about 1,600 U.S. places, 4 years from 2010 to 2013, where data points are collected every 1 hour with each data point consisting of the target value “wet bulb” and 11 climate features. 

We run the forecasting tasks about prediction length of 48 and 1440 on ETT dataset and visualize the forecasting results of BTSF, TNC and supervised models. From Figure \ref{fig:foree} and \ref{fig:foree2}, we could find that our BTSF achieves the best forecasting results under both short-term and long-term settings since it adequately leverages the global context and utilize temporal-spectral relations which are helpful in producing more accurate predictive representations. 

\subsection{Anomaly detection}
In Section \ref{experiment}, we conduct extensive experiments about time series anomaly detection on five widely used datasets, which are all public available. The detailed introduction to these datasets are illustrated as follows:
\paragraph{Secure Water Treatment (SWaT)}
The SWaT dataset is a scaled down version of a real-world industrial water treatment plant producing filtered water \citep{goh2016dataset}. The collected dataset \citep{mathur2016swat} consists of 11 days of continuous operation: 7 days collected under normal operations and 4 days collected with attack scenarios.
\paragraph{Water Distribution (WADI)}
This dataset is collected from an extension of the SWaT tesbed. It consists of 16 days of continuous operation: 14 days were collected under normal operation and 2 days with attack scenarios.
\paragraph{Server Machine Dataset (SMD)}
This dataset is a 5-week-long dataset from a large internet company which was collected and made publicly available \citep{su2019robust}. It contains data from 28 server machines with each one monitored by m=33 metrics. SMD is divided into two subsets of equal size: the first half is the training set and the second half is the testing set.
\paragraph{Soil Moisture Active Passive (SMAP) and Mars Science Laboratory (MSL)}
SMAP and MSL are two real-world public datasets, expert-labeled datasets from NASA \citep{hundman2018detecting}.

\end{document}